\documentclass[12pt]{article}
\usepackage{arxiv}
\usepackage[utf8]{inputenc} % allow utf-8 input
\usepackage[T1]{fontenc}    % use 8-bit T1 fonts
\usepackage{cite}
\usepackage[colorlinks,bookmarksopen,bookmarksnumbered,citecolor=blue,urlcolor=black]{hyperref}
\usepackage{url}            % simple URL typesetting
\usepackage{booktabs}       % professional-quality tables
\usepackage{amsfonts}       % blackboard math symbols
\usepackage{nicefrac}       % compact symbols for 1/2, etc.
\usepackage{microtype}      % microtypography
\usepackage{graphicx}
\usepackage{doi}

\usepackage{caption}
\usepackage{subcaption}

\makeatletter
\renewcommand\normalsize{\@setfontsize\normalsize{12pt}{12pt}}
\makeatother

\title{Kinematically Controllable Cable Robots with Reconfigurable End-effectors}

\author{ \href{https://orcid.org/0000-0003-2891-0658}{\includegraphics[scale=0.06]{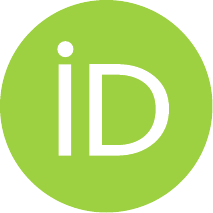}\hspace{1mm}Nan Zhang}
    \\
	\texttt{robotnanzhang@gmail.com} 
}

\renewcommand{\headeright}{Technical Brief}
\renewcommand{\undertitle}{Technical Brief}
\renewcommand{\shorttitle}{}

\hypersetup{
pdftitle={Kinematically Controllable Cable Robots with Reconfigurable End-effectors},
pdfsubject={q-bio.NC, q-bio.QM},
pdfauthor={Nan Zhang},
pdfkeywords={Cable robots},
}

\begin{document}
\maketitle

\begin{abstract}
	To enlarge the translational workspace of cable-driven robots, one common approach is to increase the number of cables. However, this introduces two challenges: (1) cable interference significantly reduces the rotational workspace, and (2) the solution of tensions in cables becomes non-unique, resulting in difficulties for kinematic control of the robot. In this work, we design structurally simple reconfigurable end-effectors for cable robots. By incorporating springs, a helical-grooved shaft, and a matching nut, relative linear motions between end-effector components are converted into relative rotations and grasping motions, thereby expanding the rotational workspace of the mechanism and enabling object grasping without actuators mounted on the end-effector. In particular, a bearing is introduced to provide an additional rotational degree of freedom, making the mechanism non-redundant. As a result, the robot's motion can be controlled purely through kinematics without additional tension sensing and control. Possible applications of cable robots designed based on this concept include large workspace motion simulation and underwater pick-and-place operations.
\end{abstract}

% keywords can be removed

\section{Introduction}
Cable robots drive their end-effectors via cables to achieve desired tasks. Owing to their large translational workspace and high payload-to-weight ratio, they have been used in crane-type operations \cite{albus1992nist}, astronomical observatories \cite{roshi2021future}, and immersive virtual reality \cite{miermeister2016cablerobot}.

Typical cable-driven robots have an equal number of actuators and degrees of freedom (DoFs), and their structures are fixed. Systems whose end-effectors or bases can be structurally altered are categorized as reconfigurable cable-driven robots. Depending on the relationship between actuator number and system DoFs, cable robots can be classified as redundantly actuated (the number of DoFs is less than the number of actuators) or non-redundantly actuated (the number of DoFs is not less than the number of actuators).

% %
% \begin{figure}[htbp]
%     \centering
%     \includegraphics[width=0.5\textwidth]{Figs/delta1.jpeg}
%     \caption{delta}
%     \label{fig:gripper}
% \end{figure}
% %

\begin{figure}[htbp]
\centering
\begin{subfigure}{0.4\textwidth}
    \centering
    \includegraphics[width=\textwidth,height=0.3\textheight]{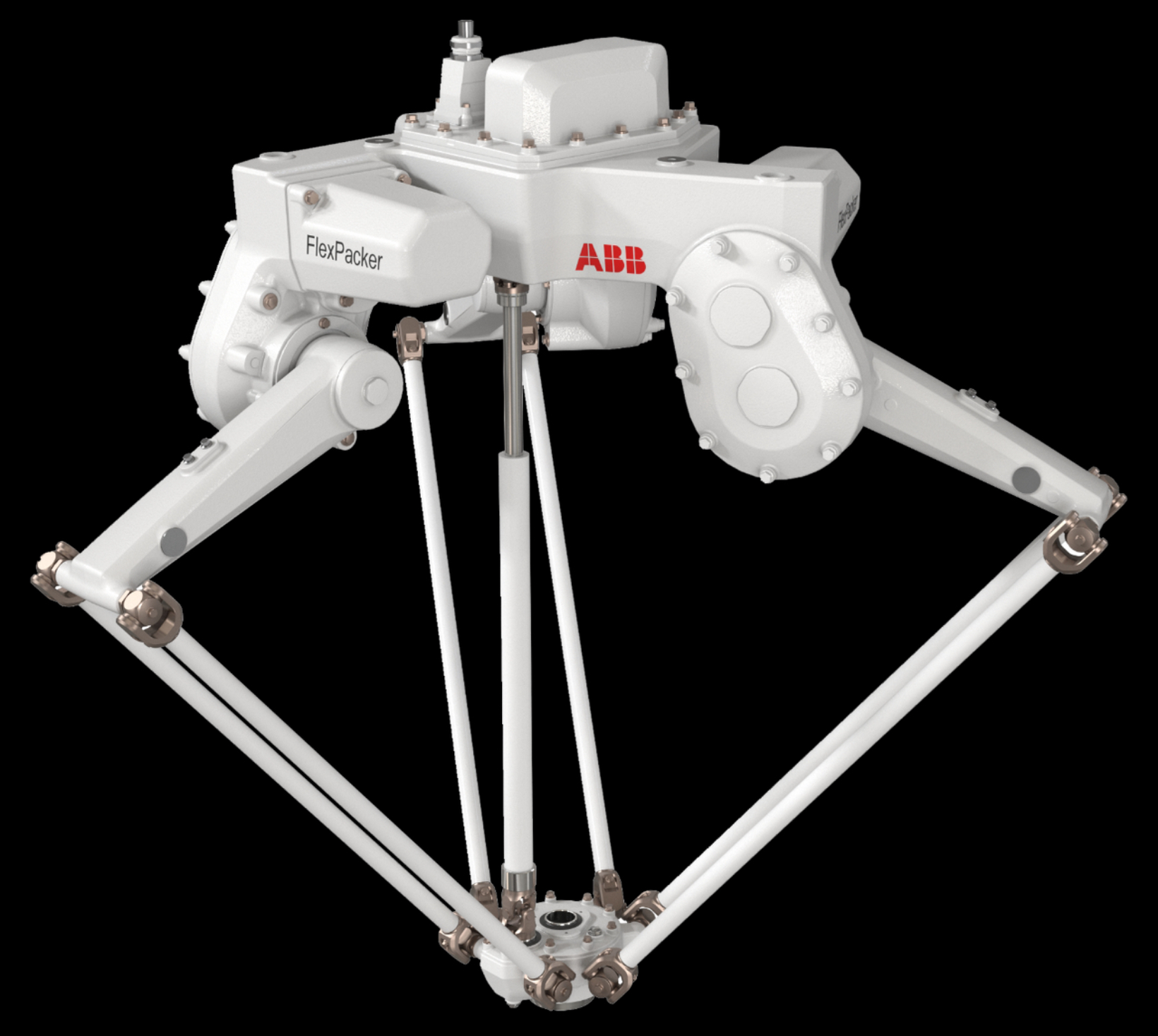}
    \caption{}
    \label{fig:a}
\end{subfigure}
\hspace*{2em}
\begin{subfigure}{0.4\textwidth}
    \centering
    \includegraphics[width=\textwidth,height=0.3\textheight]{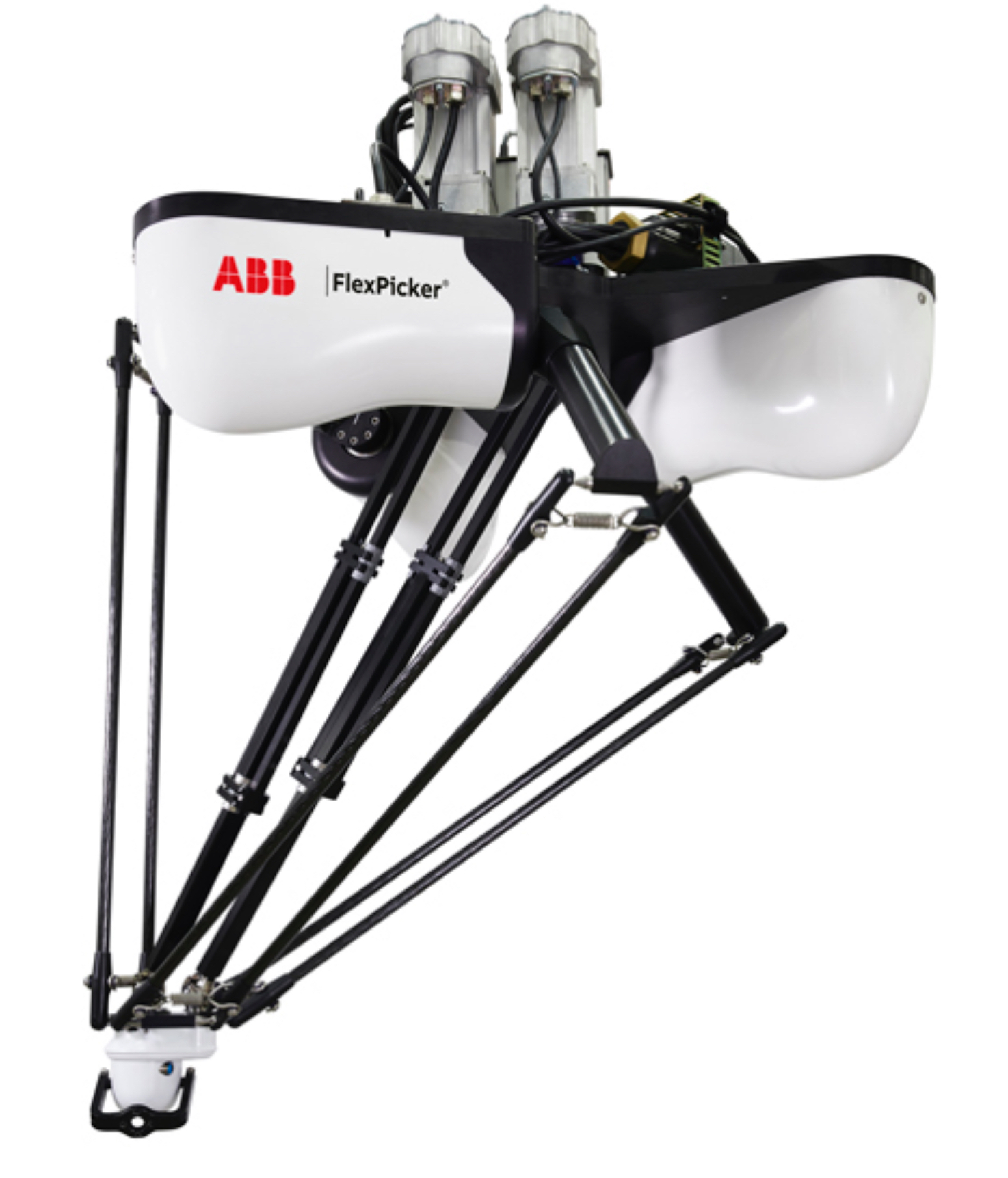}
    \caption{}
    \label{fig:b}
\end{subfigure}
\caption{Delta robot with reconfigurable end-effectors}
\label{delta}
\end{figure}

Non-reconfigurable and non-redundant cable-driven robots are commonly found in the literature. However, since cables can only transmit tensile forces, fast control of the end-effector becomes challenging. To cope with this issue, certain works \cite{behzadipour2006time} induce pneumatic mechanisms to impose pushing forces to the end-effector, while others \cite{behzadipour2006cable,behzadipour2005ultra} introduce elastic elements such as springs to maintain cable tensions. However, the rigid connection between the end effector and the base significantly limits the translational workspace, and the lack of end-effector reconfigurability results in a relatively small rotational workspace \cite{wang2022suspended}.

Traditional parallel robots face similar limitations in rotational workspace. This is typically mitigated by adding additional motor controlled links to reconfigure the end-effector, as seen in four-DoF or five-DoF architectures shown in Fig.\ref{delta}, enabling large or even unrestricted rotational motions \cite{ABB_Delta_Video_2025,ABB_IRB390_2025}. Such methods can be adapted to cable-driven robots. For example, prior work \cite{fortin2017design} introduced a drum-based end-effector driven by two additional motors, enabling rotational motion by winding cables. However, this solution is imprecise, and the rotation range is limited by cable length, though still larger than designs based on mechanical amplification using gears \cite{wang2022suspended}. Another design \cite{pott2017workspace} uses three cables to manipulate a crank mounted on the end-effector, enabling relative rotation without limits. This design is mechanically complex and introduces cable redundancy like the designs \cite{fortin2017design, wang2022suspended}, requiring tension distribution and tension sensors for dynamic control \cite{ueland2020optimal}.

To enlarge the rotational workspace of cable-driven robots, we introduce a reconfigurable end-effector that possesses two relative degrees of freedom, namely translational and rotational motions between its upper and lower components. The translational motion is converted into a wide range of rotational motion by using a spring, a helical-grooved shaft, and a matching spline nut. The rotational DoF is introduced via a bearing, ensuring that the overall mechanism remains non-redundant. Consequently, the system can be controlled purely through the control of cable lengths, without the need for expensive cable tension or motor torque sensors. Unlike prior designs, the mechanism does not impose any rigid component between the payload and the base. It relies solely on cable connections, allowing a larger translational motion range compared to \cite{behzadipour2005ultra,behzadipour2006cable} while retaining the ability to generate pushing effects, helping to maintain cable tensions and increasing operation speed.

\section{Mechanism Design}\label{sec:headings}

In this section, we first design two cable-driven robots and briefly analyze their properties. We then identify their limitations and discuss possible improvements. Finally, we present several potential applications for each design.

\subsection{Design of a Reconfigurable Cable Robot with Large Translational and Rotational Workspaces}

\begin{figure}[htbp]
\centering
\begin{subfigure}{0.44\textwidth}
    \centering
    \includegraphics[width=\textwidth,height=0.35\textheight]{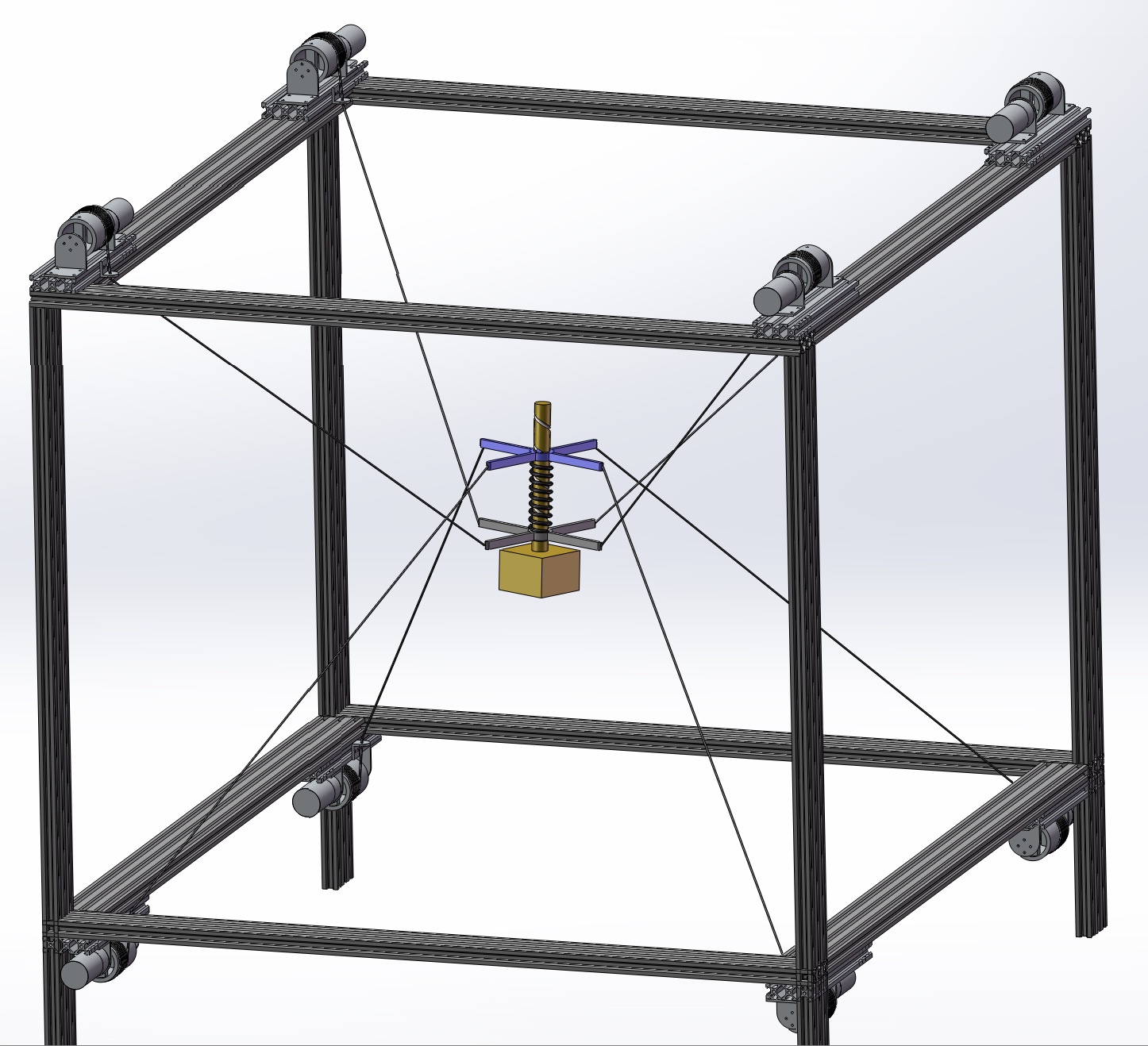}
    \caption{}
    \label{fig:a}
\end{subfigure}
\hspace*{2em}
\begin{subfigure}{0.44\textwidth}
    \centering
    \includegraphics[width=\textwidth,height=0.35\textheight]{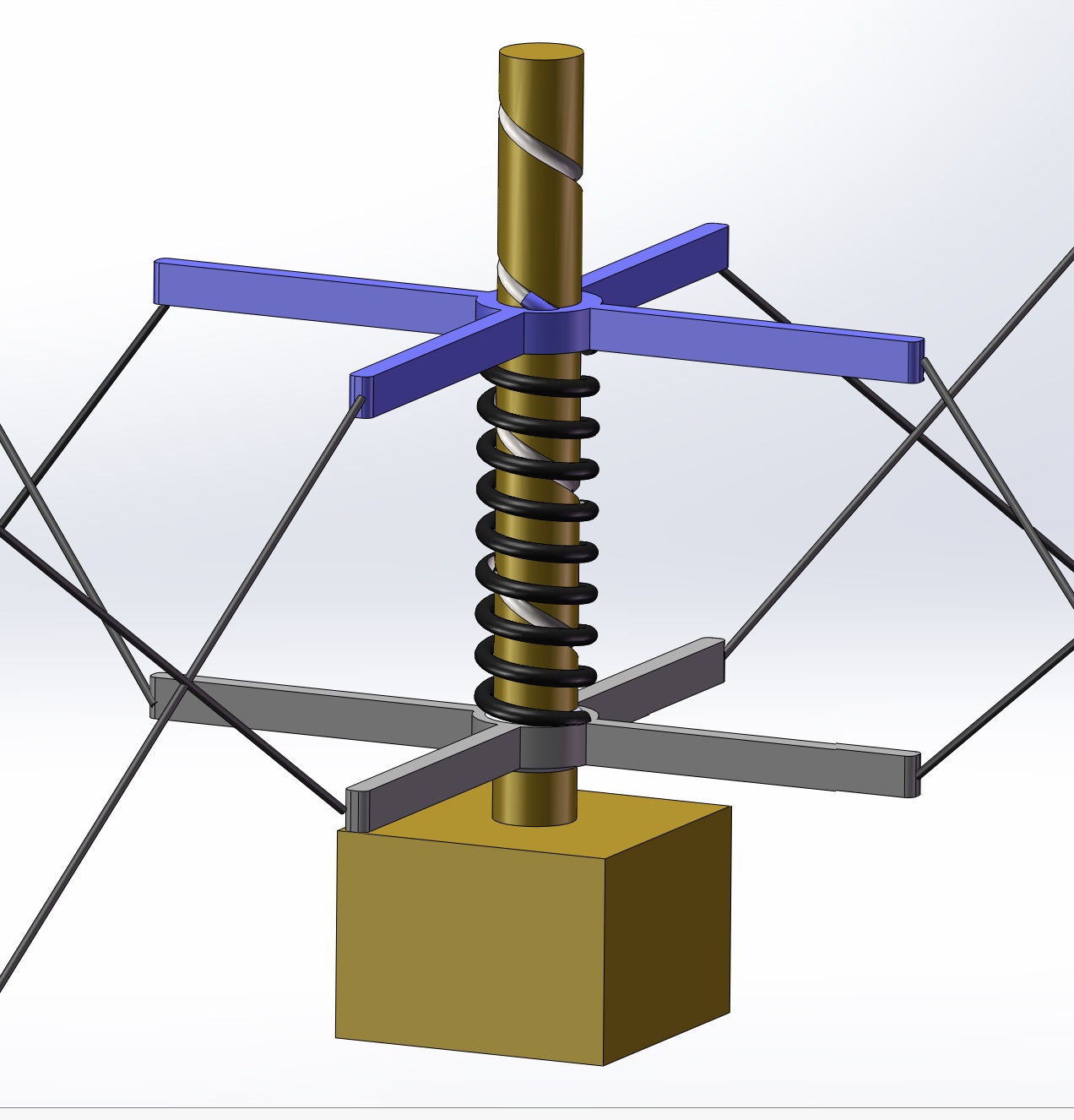}
    \caption{}
    \label{fig:b}
\end{subfigure}
\caption{Cable robot with a reconfigurable end-effector
}
\label{cablerobot1}
\end{figure}

\begin{figure}[htbp]
     \centering
     \includegraphics[width=0.4\textwidth]{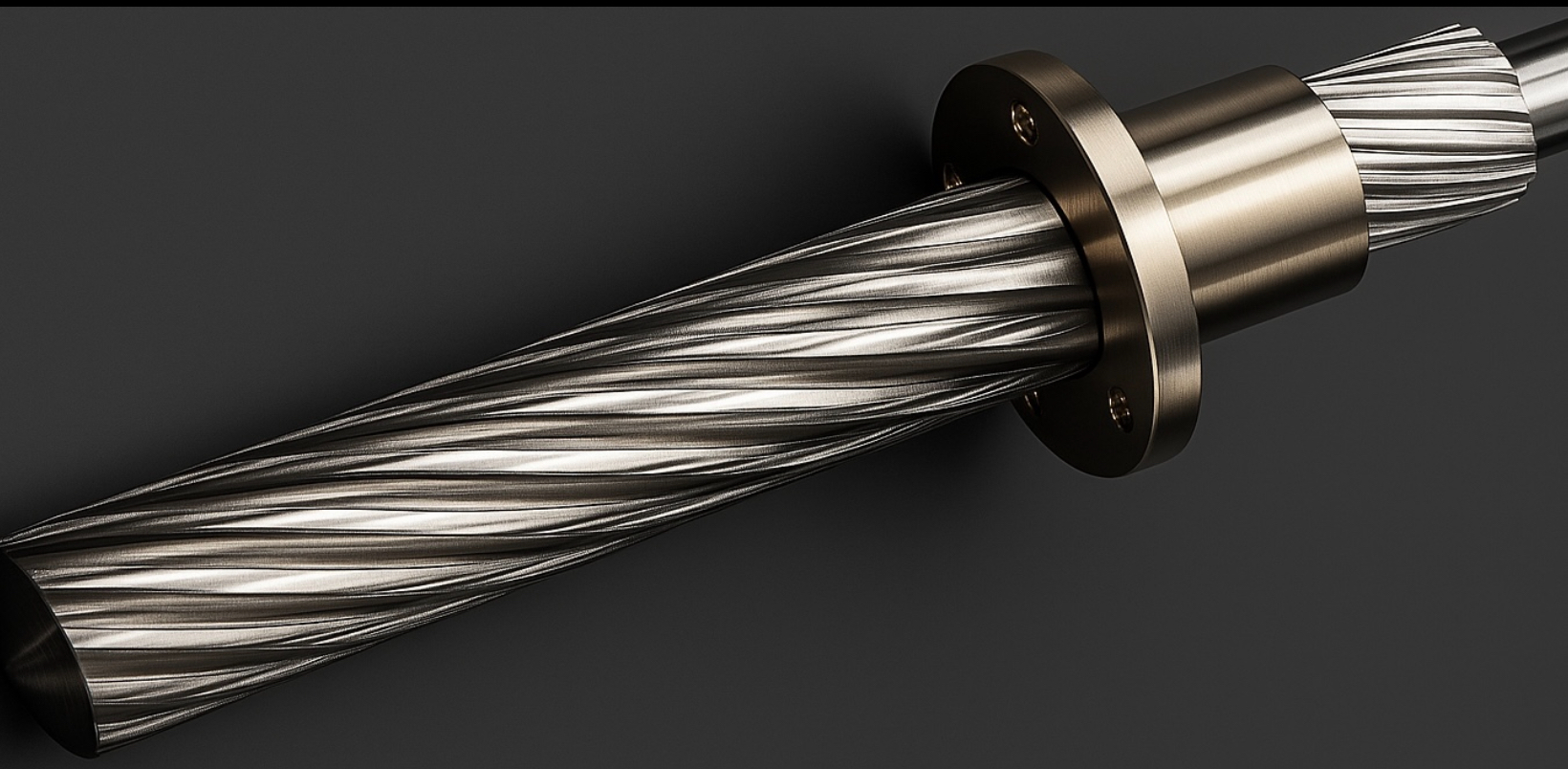}
     \caption{Helical spline shaft and matching spline nut}\label{screw}
\end{figure}

We have previously demonstrated that the reconfiguration of cable robots' anchors can be achieved solely through cables \cite{zhang2017design}. Here, we similarly accomplish end-effector reconfiguration using cables. The designed cable robot is shown in Fig.\ref{cablerobot1}, where eight cables come out from servo-motor-driven reels and connect to an end-effector possessing two degrees of freedom for relative motion. This end-effector is composed of an upper part (colored blue), a lower part (colored gray), a compression spring, and a helical-grooved shaft. The upper part contains a matching spline nut, while the lower part houses a bearing. The bearing is fixed to the shaft, allowing the lower component to rotate freely relative to it, while the nut can slide axially, thereby providing relative translational and rotational freedom. When the spring is compressed by controlling cable lengths, the shaft (and the attached cubic-shaped payload) will rotate. The design insight is motivated by spin-mop mechanisms, which convert linear actuation into rotational motion, see Fig.\ref{screw}. Unlike spin-mops, the design avoids ratchet components to ensure bidirectional rotation. Since cable forces are inherently unidirectional, the relative motion between the two parts cannot self-recover once displacement occurs. Therefore, the spring is essential for restoring the mechanism’s configuration. The workspace of the proposed mechanism exhibits geometric symmetry, thereby enabling omnidirectional rotational capability. Additionally, akin to traditional redundant manipulators, it inherits a substantially large translational workspace.

\begin{figure}[htbp]
\centering
\begin{subfigure}{0.25\textwidth}
    \centering
\includegraphics[width=\linewidth]{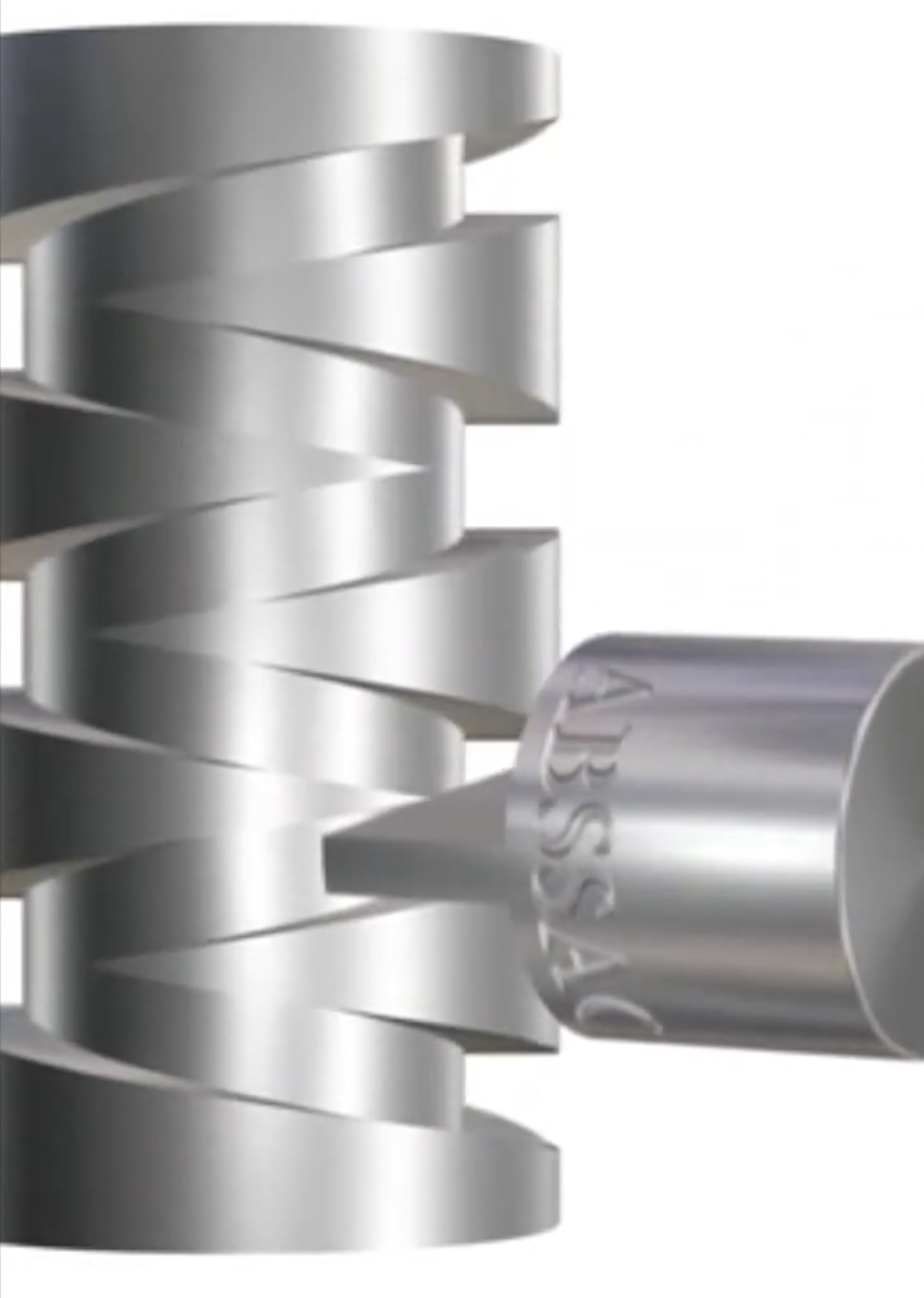}
    \caption{}\label{fig:a}
\end{subfigure}
\begin{subfigure}{0.25\textwidth}
    \centering
\includegraphics[width=\linewidth]{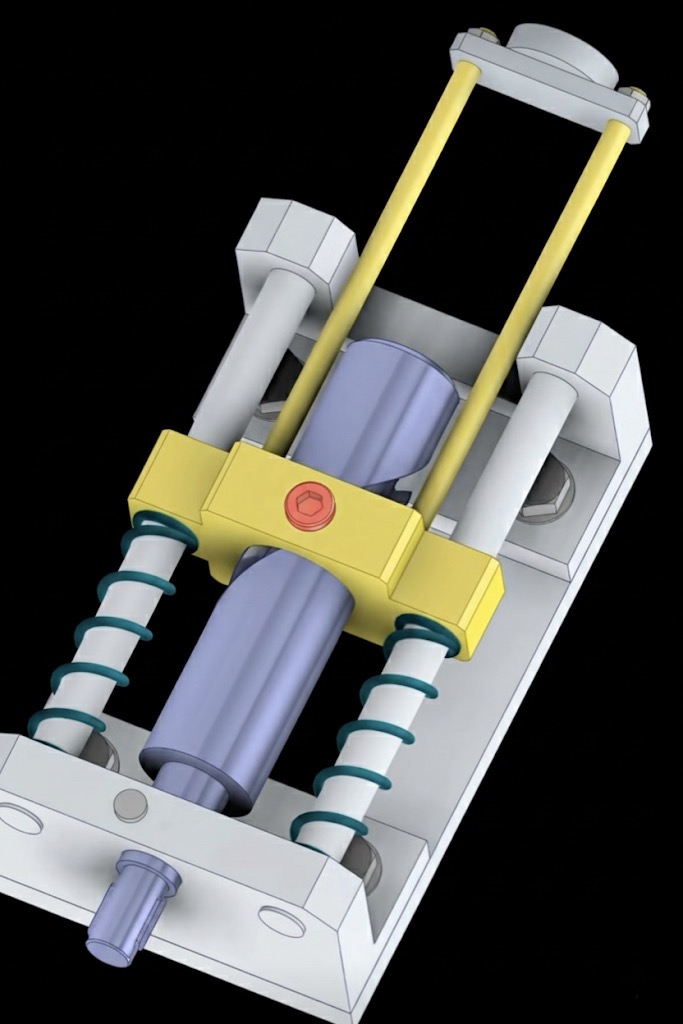}
    \caption{}\label{fig:a}
\end{subfigure}
\begin{subfigure}{0.25\textwidth}
    \centering
\hspace*{2em}
\includegraphics[width=\linewidth]{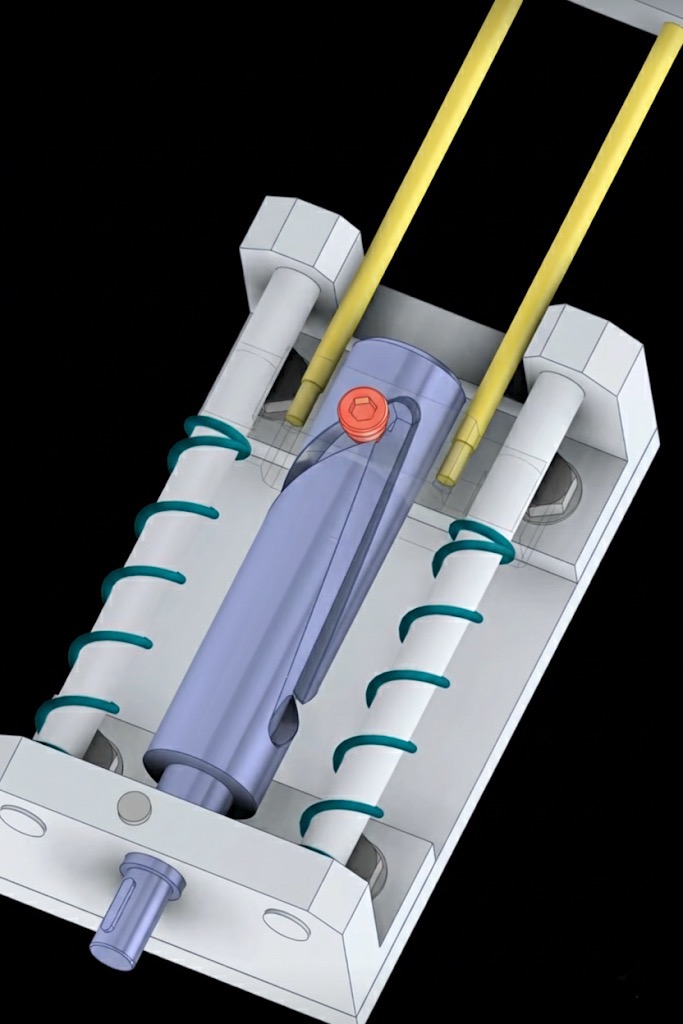}
    \caption{}\label{fig:b}
\end{subfigure}
\caption{A self-reversing winder-like end-effector}
\label{full}
\end{figure}

Nevertheless, the mechanism remains non-redundant, as the bearing-based rotation in the lower part introduces an extra relative rotational DoF (This cannot be achieved by using the design shown in Fig.\ref{full}). The end-effector inherently possesses six DoF, and the inclusion of two additional relative DoF yields a total of eight, which is precisely matched by the eight driving cables. As a result, the mechanism is classified as non-redundant. 

However, similar to redundant cable-driven architectures, it can still generate thrust-like effects, thereby enlarging the translational workspace and enhancing the velocity performance of the end-effector. Furthermore, owing to the intrinsic coupling between cable tensions and cable lengths (wherein each cable is indispensable \cite{zhang2025design} for maintaining the kinematic states of the end-effector), the motion of the end-effector can be effectively realized through purely kinematic control. In other words, the desired motions can be achieved by regulating cable lengths alone while ensuring proper tensions whose magnitude may be optimized with respect to safety and energy criteria, without relying on tensioning devices \cite{zhang2025design} or sophisticated tension control schemes \cite{ueland2020optimal}.

Both the upper and lower parts are designed with four outwardly protruding rods to reduce potential cable-end-effector interference and to decrease the sensitivity of rotational errors with respect to cable length variations. In other words, the same rotational error corresponds to larger cable length deviations, which improves the conditional numbers of the Jacobian matrix and facilitates easier kinematic control.

\begin{figure}[htbp]
\centering
\begin{subfigure}{0.44\textwidth}
\centering
\includegraphics[width=\textwidth,height=0.35\textheight]{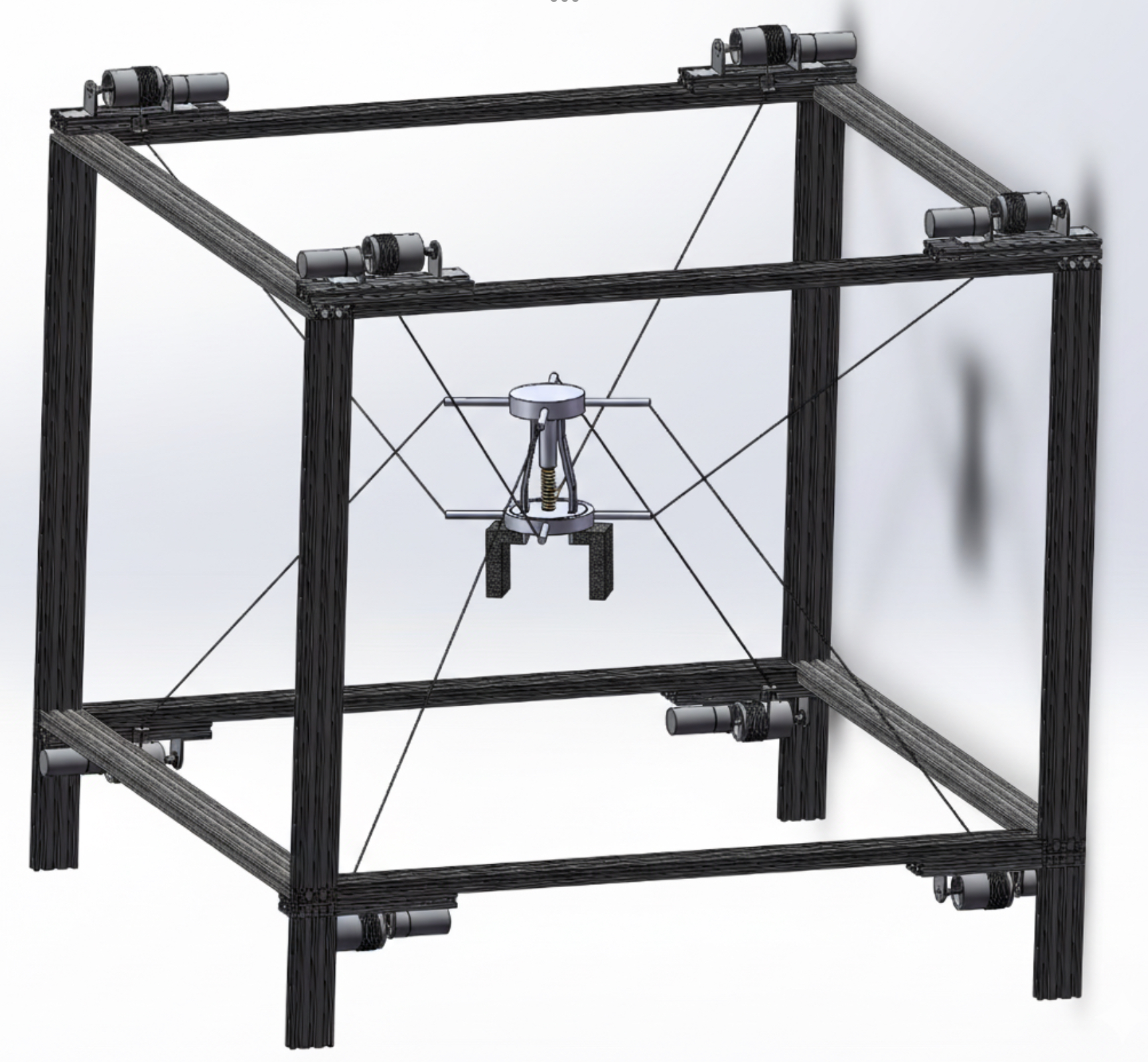}
    \caption{}
    \label{fig:a}
\end{subfigure}
\hspace*{2em}
\begin{subfigure}{0.44\textwidth}
\centering
\includegraphics[width=\textwidth,height=0.35\textheight]{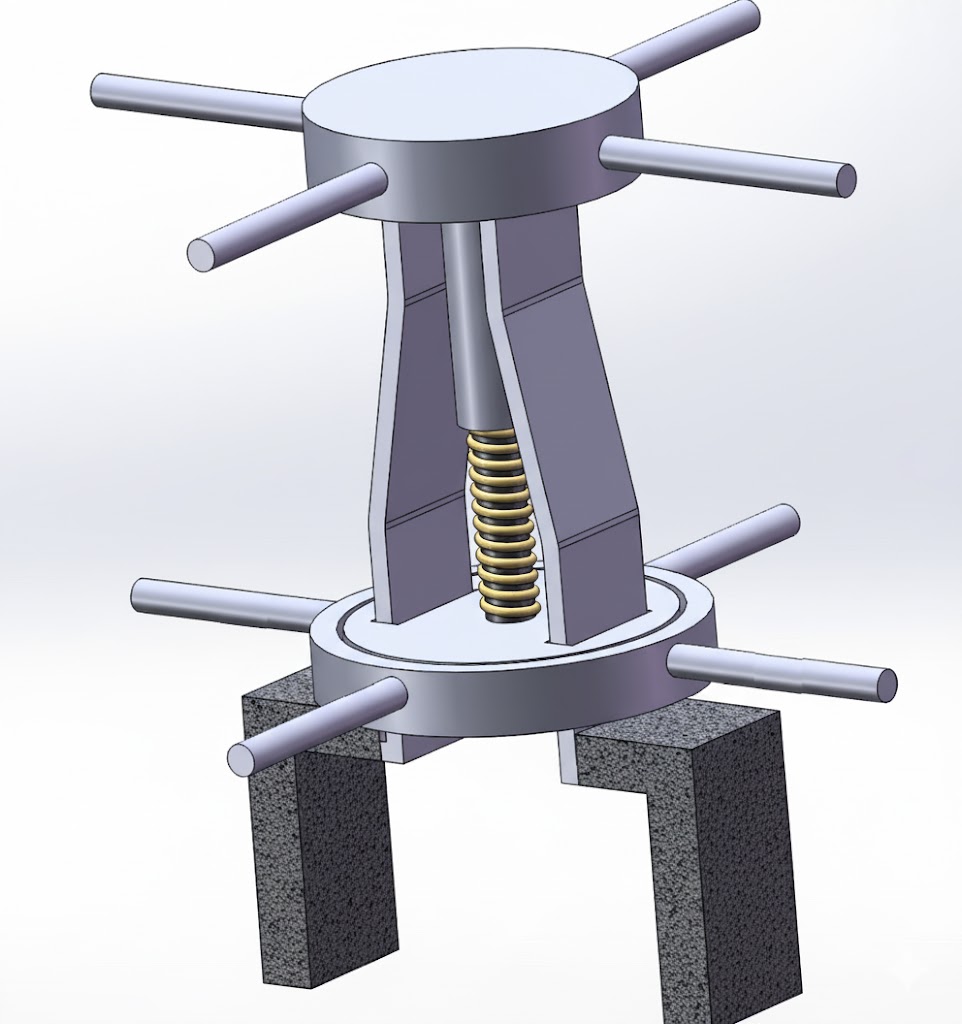}
    \caption{}
    \label{fig:b}
\end{subfigure}
\caption{A large space cable-driven gripper}
\label{cablerobot2}
\end{figure}

Although the proposed mechanism effectively resolves cable-cable and cable-end-effector interference, cable-environment interference must also be considered. By swapping the positions of the upper and lower components (as shown in the alternative design in Fig.\ref{cablerobot2}), the payload-supporting rods extend downward, spatially separating the cables from the environment contact region. This configuration is more suitable when the payload interacts with its surroundings. Another technique for avoiding cable–environment interference is to place all cable actuation units on top of the robot frame so that all cable anchor points lie on a single horizontal plane \cite{kawamura1995textordfemininedevelopment}. Although this configuration limits the robot’s workspace along the vertical axis, it enables the system to perform grinding, polishing, and cleaning operations—tasks that require hybrid position–force control. In such applications, approximate force control along the shaft can be achieved by adjusting the spring length. In this case, it is preferable to replace the helical spline shaft with a D-shaped shaft, as rotation along the shaft is unnecessary.

The end-effector in this design possesses a relative degree of freedom in the vertical direction. Likewise, it can be configured to have a relative degree of freedom in the horizontal direction, which can then be transformed into relative rotation of the end-effector via a gear–rack mechanism. These designs can rotate much larger than 360 degrees, but their rotation angles are still restricted. If unlimited rotation is required, a simple and practical solution is shown in Fig.\ref{full}. This design adopts a self-reversing winder-like shaft. By appropriately controlling the spring compression, the end-effector can achieve continuous rotation in both directions. However, this end-effector suffers from two singular configurations, making high-speed rotational control challenging. In practical applications, auxiliary sensors such as laser rangefinders may be integrated to identify the singular configurations to facilitate their smooth traversal. By complementing rotary encoders, which often provide imprecise cable length measurements, the overall control accuracy of the system may be improved.

One potential application of this mechanism is a low-cost virtual reality (VR) platform that simulates flying or free-fall scenarios. However, since this design cannot physically grasp objects, its application range remains limited.

\subsection{Design of a Cable-Driven Gripper for Large-Workspace Operations}

The design of the cable-driven gripper is illustrated in Fig.\ref{cablerobot2}. Similar to the previous design, the end-effector incorporates relative translational and rotational degrees of freedom. However, in this design, the bearing is placed at the bottom, and an elastic gripper is introduced. During the compression and release of the yellow spring, the opening width of the soft gripper changes accordingly. This gripper adopts a normally-closed configuration, indicating that it does not require additional power to maintain holding force, thereby reducing energy consumption.

\begin{figure}[htbp]
\centering
\begin{subfigure}{0.2\textwidth}
    \centering
    \includegraphics[width=\textwidth,height=0.2\textheight]{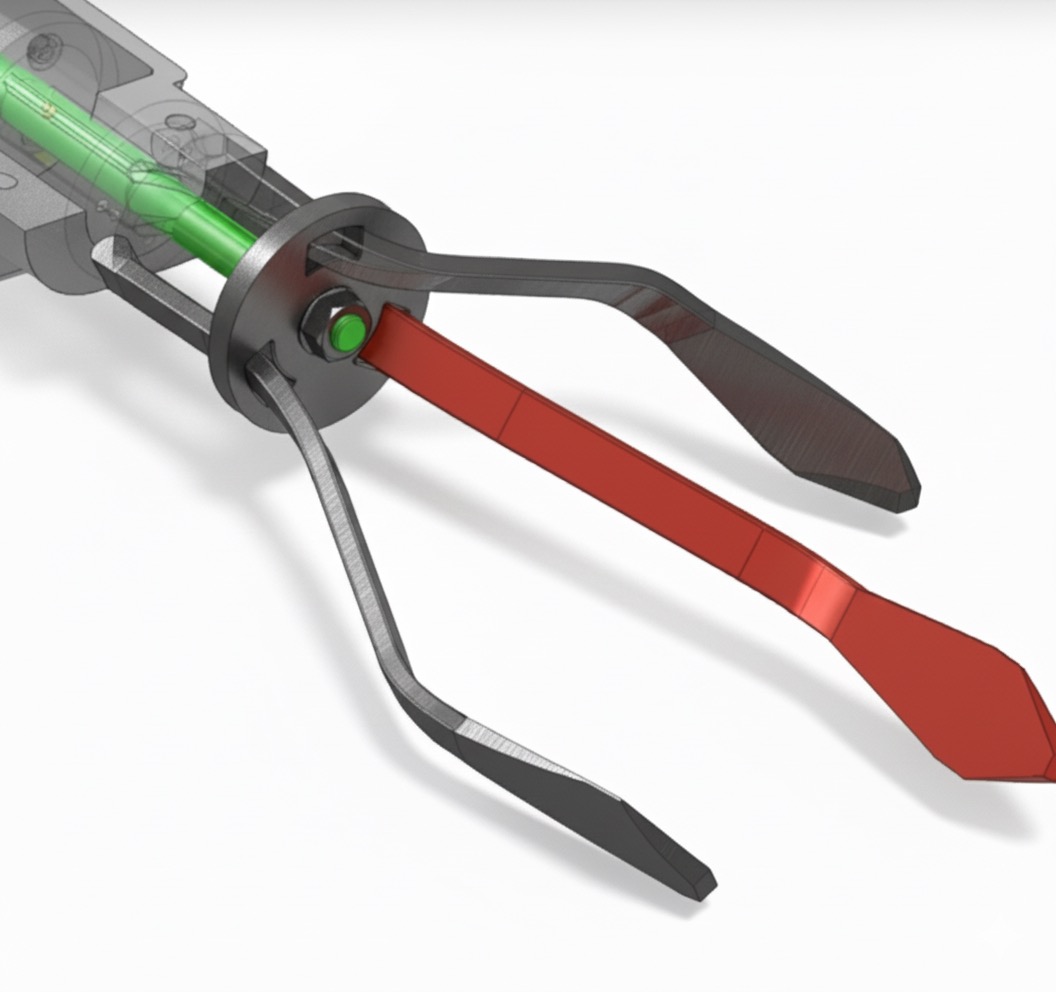}
    \caption{}
    \label{fig:b}
\end{subfigure}
\hfill
\begin{subfigure}{0.2\textwidth}
    \centering
    \includegraphics[width=\textwidth,height=0.2\textheight]{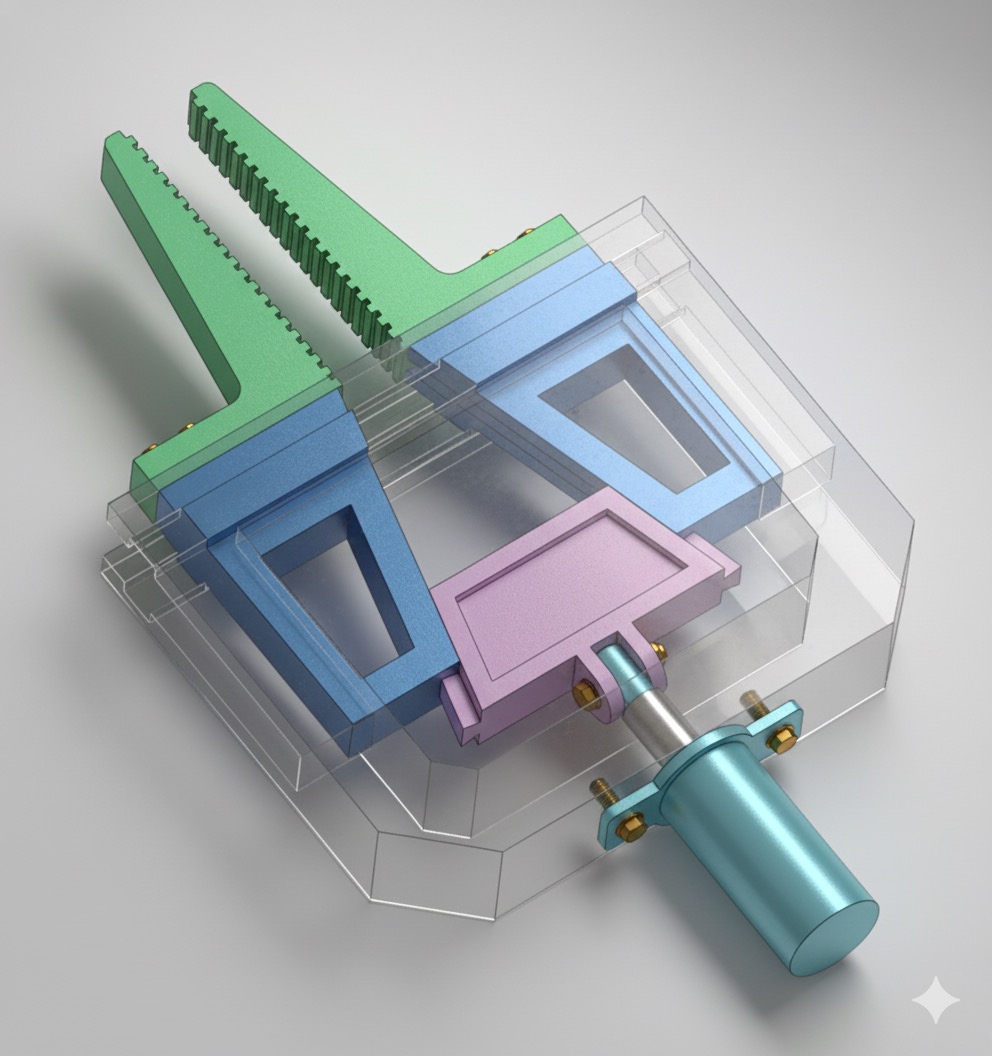}
    \caption{}
    \label{fig:a}
\end{subfigure}
\hfill
\begin{subfigure}{0.2\textwidth}
    \centering
    \includegraphics[width=\textwidth,height=0.2\textheight]{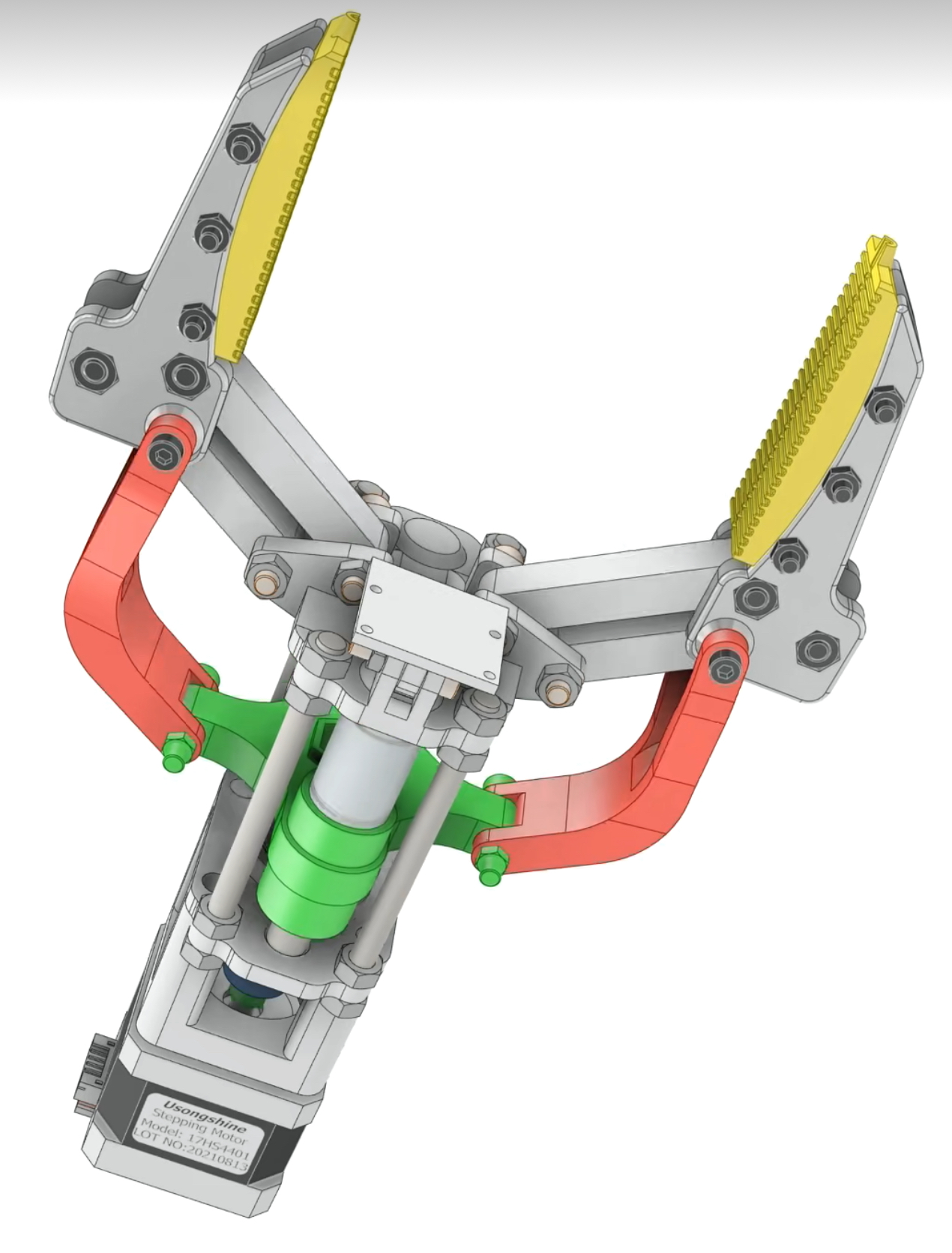}
    \caption{}
    \label{fig:b}
\end{subfigure}
\hfill
\begin{subfigure}{0.2\textwidth}
    \centering
    \includegraphics[width=\textwidth,height=0.2\textheight]{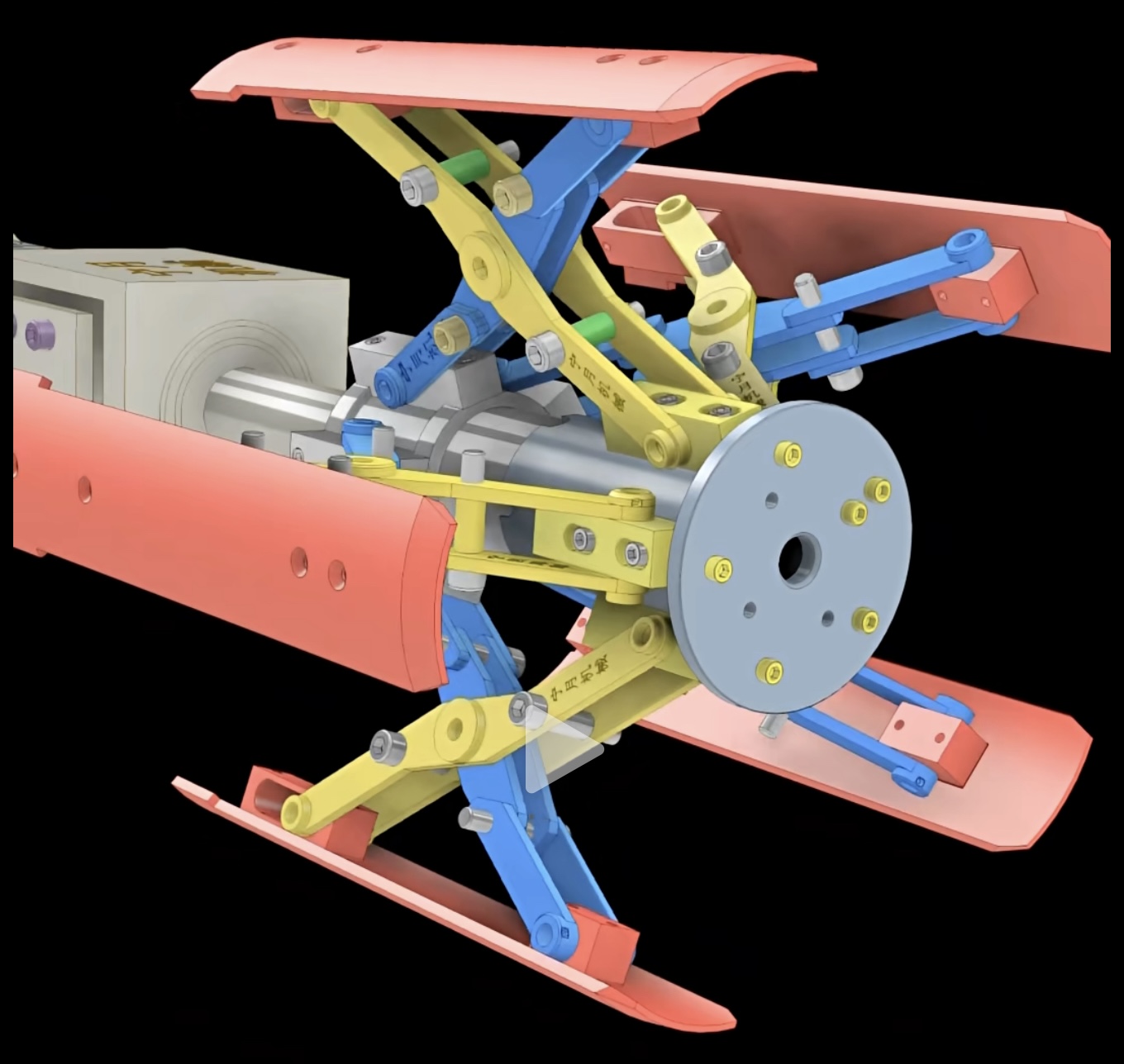}
    \caption{}
    \label{fig:b}
\end{subfigure}
\caption{Other types of grippers that are suitable for remote control by cables}
\label{grippers}
\end{figure}

It is worth noting that when the gripper is fully closed, small variations in spring length do not significantly affect the grasping width. This design feature helps reduce the risk of grasping failure caused by spring-length control errors. We can also use cables to manipulate other practical grippers (see Fig.\ref{grippers} with the help of springs and bearings. To accommodate objects of varying shapes and sizes, these grippers can be designed with a large opening/closing range.

The design is suitable for pick-and-place tasks involving objects that are not too heavy. However, the mechanism's rotational workspace is too small compared to the mechanism designed before. We can incorporate a rack-and-pinion system or a planetary reducer to enable simultaneous gripping and wide range rotational motions, but the resulting design will become much more complex.

\subsection{Design of a Rotatable Cable-Driven Gripper for Large-Workspace Pick-and-Place Operations}

\begin{figure}[htbp]
\centering
\begin{subfigure}{0.44\textwidth}
\centering
\includegraphics[width=\textwidth,height=0.35\textheight]{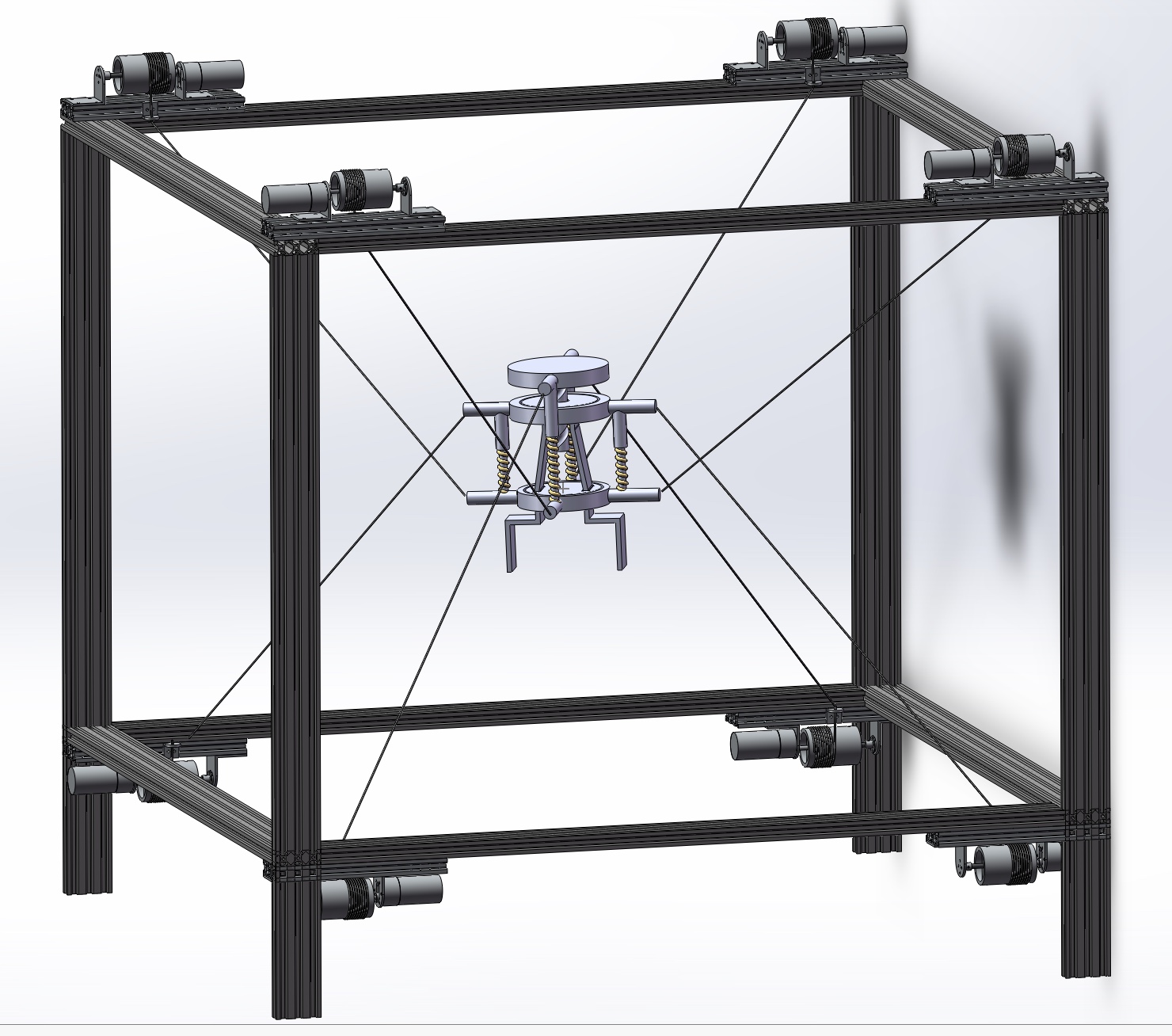}
    \caption{}
    \label{fig:a}
\end{subfigure}
\hspace*{2em}
\begin{subfigure}{0.44\textwidth}
\centering
\includegraphics[width=\textwidth,height=0.35\textheight]{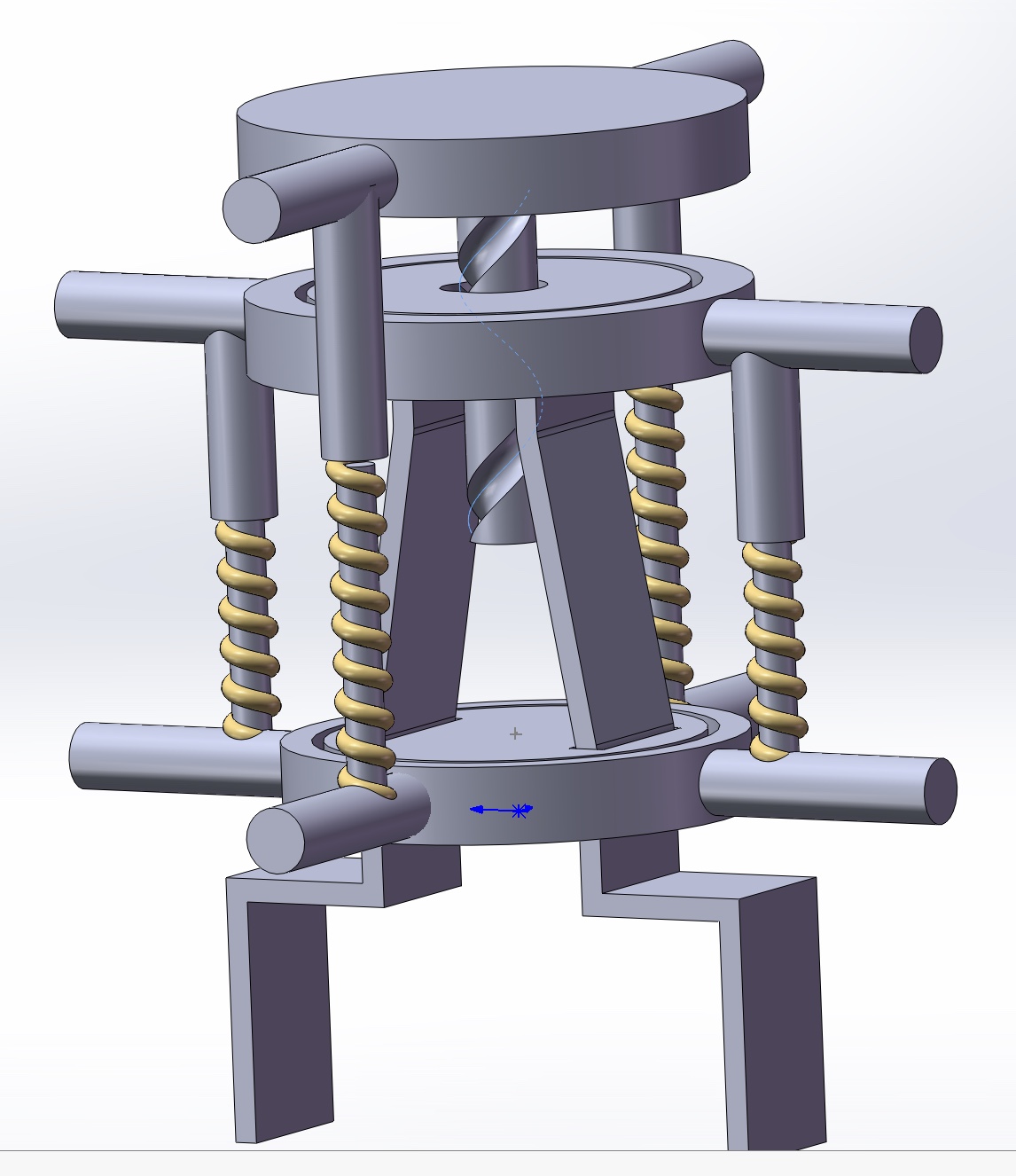}
    \caption{}
    \label{fig:b}
\end{subfigure}
\caption{A cable-driven gripper with a fully rotational workspace}
\label{cablerobot3}
\end{figure}

To achieve 360-degree rotation and grasping, the end-effector can be designed with two relative translational degrees of freedom. One controls the translational motion of the gripper, while the other governs its rotation. Unlike the previous designs, the proposed gripper consists of three parts—the upper part, middle part, and lower part—as shown in Fig.\ref{cablerobot3}. Two shafts and two springs are installed between the upper and lower parts, and their relative motion controls the opening and closing of the gripper. A similar structure is implemented between the upper and middle parts, but with an additional helical-grooved shaft fixed to the upper part and a matching spline nut fixed to the middle part. This configuration converts their relative translational motion into relative rotational motion, thereby realizing the gripper’s rotational capability. Bearings are mounted on both the middle and lower parts to prevent cable interference during rotation. It should be noted that, unlike the design in Fig.\ref{cablerobot1}, these bearings do not introduce additional degrees of freedom, as their rotations are entirely determined by the two translational motions. Consequently, the proposed mechanism has an equal number of degrees of freedom and cables, allowing for straightforward kinematic control. By mounting two laser sensors on the end-effector to measure the relative displacements of the three parts, the grasping force and angle may be reliably controlled.

Considering the actuation is realized by using cables instead of motors mounted on the end-effector \cite{lin2018design}, the gripper can also be used in underwater operations. In fact, by relocating the four lower actuators to the upper frame (while keeping their pulley positions unchanged), the entire gripper assembly can be submerged, while all actuators remain above the water surface. Since no powered actuators are located in the water, the system can operate safely and reliably in submerged environments.

The dynamic equations of the proposed mechanisms can be derived using Newton–Euler or Lagrangian formulations. Although dynamic models and optimization-based tension solvers should be employed for fast and  efficient manipulation, they are used solely to compute the desired cable lengths (or the desired end-effector configurations), which serve as inputs to the joint-space (or configuration-space) kinematic controller \cite{lynch2017modern}. Based on these models, static and dynamic workspace analysis, stiffness evaluation/control, parameter optimization (such as spring stiffness and length), motion planning, and controller design using feedback information can be systematically conducted \cite{zhang2021motion}.

\section{Conclusion}
In this work, we presented two simple cable-driven robots whose end-effectors can be passively reconfigured through the use of springs. Although the number of cables exceeds the task-required degrees of freedom, the overall system remains non-redundant because additional relative degrees of freedom are introduced at the end-effector, matching the total system degrees of freedom to the number of cables. As a result, cable tensions can be satisfied through purely kinematic control, avoiding the need for actuators to operate in torque control mode, which is typically difficult to implement in practice. The passive reconfiguration of the end-effector primarily operates through spring compression and recovery. The helical spline integrated at the end-effector converts the spring-induced linear displacement into relative rotational motion, mitigating cable interference and slackness during large or even unlimited rotational movements. Owing to its symmetric architecture and easy-to-reconfigure property, the proposed design provides a large translational workspace and can accomplish non-contact/contact tasks such as grinding/polishing/cleaning, virtual reality that simulates flying scenarios and grasping underwater objects.

\bibliographystyle{unsrt}
\bibliography{references}  

\begin{thebibliography}{10}

\bibitem{albus1992nist}
James Albus, Roger Bostelman, and Nicholas Dagalakis.
\newblock The nist spider, a robot crane.
\newblock {\em Journal of research of the National Institute of Standards and Technology}, 97(3):373--385, 1992.

\bibitem{roshi2021future}
D~Anish Roshi, N~Aponte, E~Araya, H~Arce, LA~Baker, W~Baan, TM~Becker, JK~Breakall, RG~Brown, CGM Brum, et~al.
\newblock The future of the arecibo observatory: The next generation arecibo telescope.
\newblock {\em arXiv preprint arXiv:2103.01367}, 2021.

\bibitem{miermeister2016cablerobot}
Philipp Miermeister, Maria L{\"a}chele, Rainer Boss, Carlo Masone, Christian Schenk, Joachim Tesch, Michael Kerger, Harald Teufel, Andreas Pott, and Heinrich~H B{\"u}lthoff.
\newblock The cable robot simulator large scale motion platform based on cable robot technology.
\newblock In {\em 2016 IEEE/RSJ International Conference on Intelligent Robots and Systems (IROS)}, pages 3024--3029. IEEE, 2016.

\bibitem{behzadipour2006time}
Saeed Behzadipour and Amir Khajepour.
\newblock Time-optimal trajectory planning in cable-based manipulators.
\newblock {\em IEEE Transactions on Robotics}, 22(3):559--563, 2006.

\bibitem{behzadipour2006cable}
Saeed Behzadipour and Amir Khajepour.
\newblock {\em Cable-based robot manipulators with translational degrees of freedom}.
\newblock IntechOpen, 2006.

\bibitem{behzadipour2005ultra}
Saeed Behzadipour.
\newblock Ultra-high-speed cable-based robots.
\newblock {\em The degree of PhD thesis, University of Waterloo}, 2005.

\bibitem{wang2022suspended}
Ruobing Wang, Shufei Li, and Yangmin Li.
\newblock A suspended cable-driven parallel robot with articulated reconfigurable moving platform for sch{\"o}nflies motions.
\newblock {\em IEEE/ASME Transactions on Mechatronics}, 27(6):5173--5184, 2022.

\bibitem{ABB_Delta_Video_2025}
{ABB Robotics}.
\newblock Abb irb 390 delta robot demonstration.
\newblock \url{https://www.youtube.com/watch?v=E9YjzIlN6EI}, 2021.
\newblock YouTube video, Accessed: 2025-10-24.

\bibitem{ABB_IRB390_2025}
{Carolina Motion Controls}.
\newblock Abb irb 390: 10~kg payload with 1.3~m diameter.
\newblock \url{https://carolinamotioncontrols.com/irb-390-10-1300-10kg-payload-with-1-3m-diameter/}, 2025.
\newblock Accessed: 2025-10-24.

\bibitem{fortin2017design}
Alexis Fortin-C{\^o}t{\'e}, C{\'e}line Faure, Laurent Bouyer, Bradford~J McFadyen, Catherine Mercier, Micha{\"e}l Bonenfant, Denis Laurendeau, Philippe Cardou, and Cl{\'e}ment Gosselin.
\newblock On the design of a novel cable-driven parallel robot capable of large rotation about one axis.
\newblock In {\em Cable-Driven Parallel Robots: Proceedings of the Third International Conference on Cable-Driven Parallel Robots}, pages 390--401. Springer, 2017.

\bibitem{pott2017workspace}
Andreas Pott and Philipp Miermeister.
\newblock Workspace and interference analysis of cable-driven parallel robots with an unlimited rotation axis.
\newblock In {\em Advances in Robot Kinematics 2016}, pages 341--350. Springer, 2017.

\bibitem{ueland2020optimal}
Einar Ueland, Thomas Sauder, and Roger Skjetne.
\newblock Optimal force allocation for overconstrained cable-driven parallel robots: Continuously differentiable solutions with assessment of computational efficiency.
\newblock {\em IEEE Transactions on Robotics}, 37(2):659--666, 2020.

\bibitem{zhang2017design}
Nan Zhang, Weiwei Shang, and Shuang Cong.
\newblock Design and analysis of an under-constrained reconfigurable cable-driven parallel robot.
\newblock In {\em 2017 IEEE International Conference on Cybernetics and Intelligent Systems (CIS) and IEEE Conference on Robotics, Automation and Mechatronics (RAM)}, pages 13--18. IEEE, 2017.

\bibitem{zhang2025design}
Nan Zhang and Long Teng.
\newblock Design and motion planning of a cable robot utilizing cable slackness.
\newblock {\em Journal of Mechanisms and Robotics}, 17(5), 2025.

\bibitem{kawamura1995textordfemininedevelopment}
Sadao Kawamura et~al.
\newblock Development of an ultrahigh speed robot falcon using wire drive system.
\newblock {\em Robotics and Automation}, pages 215--220, 1995.

\bibitem{lin2018design}
Jonqlan Lin, Chi~Ying Wu, and Julian Chang.
\newblock Design and implementation of a multi-degrees-of-freedom cable-driven parallel robot with gripper.
\newblock {\em International Journal of Advanced Robotic Systems}, 15(5):1729881418803845, 2018.

\bibitem{lynch2017modern}
Kevin~M Lynch and Frank~C Park.
\newblock {\em Modern robotics}.
\newblock Cambridge University Press, 2017.

\bibitem{zhang2021motion}
Nan Zhang.
\newblock {\em Research on motion planning for cable robots [Doctoral dissertation, University of Science and Techonology of China]}.
\newblock DOI: 10.13140/RG.2.2.16880.99843, 2021.

\end{thebibliography}

\end{document}